\documentclass{article} 
\usepackage{iclr2017_workshop,times}
\usepackage{hyperref}
\usepackage{url}
\usepackage[pdftex]{graphicx}
\usepackage{amsmath}
\usepackage{amssymb}
\usepackage{esvect}

\title{Consistent Alignment of Word Embedding Models\thanks{\tiny{This material is based upon work supported by the Assistant Secretary of Defense for Research and Engineering under Air Force Contract No. FA8721-05-C-0002 and/or FA8702-15-D-0001. 
Any opinions, findings, conclusions or recommendations expressed in this material are those of the author(s) and do not necessarily reflect the views of the Assistant Secretary of Defense for Research and Engineering.}}}

\author{Cem \c{S}afak \c{S}ahin, Rajmonda S. Caceres, Brandon Oselio \& William M. Campbell  \\
MIT Lincoln Laboratory\\
244 Wood Street\\
Lexington, MA 02421, USA \\
}

\begin{document}

\maketitle

\begin{abstract}
Word embedding models offer continuous vector representations that can capture rich contextual semantics based on their word co-occurrence patterns. 
While these word vectors can provide very effective features used in many \textsc{nlp} tasks such as clustering similar words and inferring  learning relationships, many challenges and open research questions remain. 
In this paper, we propose a solution that aligns variations of the same model (or different models) in a joint low-dimensional latent space leveraging carefully generated synthetic data points. 
This generative process is inspired by the observation that a variety of linguistic relationships is captured by simple linear operations in embedded space. 
We demonstrate that our approach can lead to substantial improvements in recovering embeddings of local neighborhoods.
\end{abstract}

\section{Introduction}
\label{sec:intro}
Recently there has been a growing interest in continuous vector representations of linguistic entities, most often referred to as embeddings. 
This includes techniques that rely on matrix factorization~(\cite{omer2014,pennington2014}), as well as currently popular neural network methods~(\cite{le2014,mikolov2013efficient,mikolov2013a}). 
These embeddings are able to capture complex semantic patterns such as linguistic analogies and have shown remarkable performance improvements across various  \textsc{nlp} tasks. 

Nonetheless, continuous word representations generated by neural network models are not well understood and evaluations of these representations are still nascent. 
It is not clear what the dimensions of the word vectors represent, and as such, we are often unable to easily evaluate the quality of representation except with reference to performance of downstream tasks (e.g., clustering, domain adaptation). It is also difficult to compare word embeddings of different dimensions, and when we do this naively, we often see wildly different local properties including from models trained on the same dataset. 
Herein, we highlight some these fundamental limitations of word representations, in particular with respect to their ability to be embedded and aligned in a lower-dimensional space.



Manifold alignment techniques have a rich history of addressing high dimensionality challenges within the domain adaptation and transfer learning research areas~(\cite{wang2009general,wang2011heterogeneous}), but they have primarily been applied to data sources such as images, and genomic data.
In~\cite{wang2016a} manifold alignment techniques are used to discover \textit{logical} relationships in supervised settings. We believe that there is a great opportunity to further leverage the same techniques in unsupervised settings. However, it is not clear if these techniques will easily translate to alignment of continuous vector spaces when labels are not available.

Our main contribution consists of an approach that overcomes some of the effects of artificial high dimensionality by leveraging synthetically generated neighboring points, or as we refer to them, \textit{latent words}. 
Inspired by the surprising insight that in high dimensional space, semantically similar words relate to one another via simple linear operations~(\cite{hashimoto2016a,mikolov2013efficient,mikolov2013a}), we conjecture that unseen words and word co-occurrences in the training datasets can be imputed in the high dimensional space via simple local linear functions. 
Data imputation has been successfully used in traditional statistical analysis of incomplete datasets to improve learning and inference. 
The application of this concept however to improve the quality of word embedding representations is novel. 
As we demonstrate in Section~\ref{sec:soln}, local densification of word point clouds allows us to take much better advantage of manifold embedding and alignment techniques. 
Additionally, we can fuse and enrich different word embedding models without the need of model retraining and access to more training data.

\section{The Problem}
\label{sec:problem}
\begin{figure}
  \begin{minipage}[c]{0.47\textwidth}
    \caption{Quality of local neighborhood reservation as a function of under-sampling in the Wikipedia dataset. We consider different subsets ($100$ words and their local neighborhoods) of the data, where we vary the size of the vocabulary (represented as $|V|$) from 140K to 266K dictionary words.}
     \label{fig:wikipedia-n-overlap}
  \end{minipage}\hfill
  \begin{minipage}[c]{0.50\textwidth}
      \includegraphics[width=\textwidth]{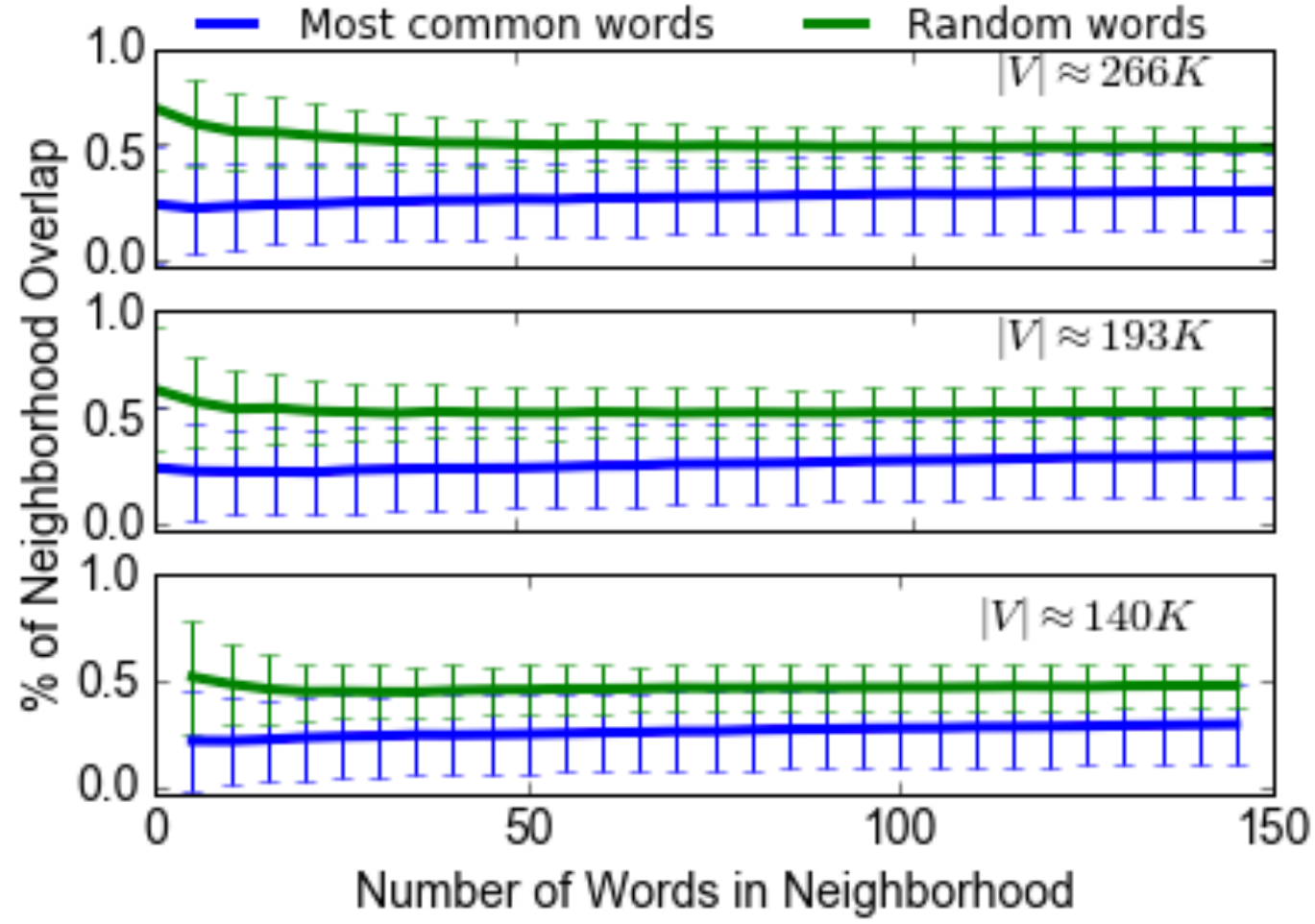}
  \end{minipage}
  \vspace{-0.5cm}
\end{figure}

We begin by illustrating how inherent randomness in the word embedding models as well as curse of dimensionality cause the output representation to be unstable and inconsistent. We report experimental results using the~\cite{wikipediaDump} dataset as presented in Fig.~\ref{fig:wikipedia-n-overlap}. 
The word embedding model is trained using the gensim library~(\cite{gensim}) on three subsets of the Wikipedia dataset, containing $140K, 193K\mbox{ and }260K$ vocabulary words. 
The smaller datasets are proper subsets of the larger ones. 
This almost doubling of vocabulary size causes only about $24\%$ increase of average frequencies of $100$ most common words. 
In all cases, the words are mapped as points in $\mathcal{R}^{200}$. 
For a fixed local neighborhood size, we re-train the same model, using the same parameters, on the same training dataset. 
We then measure model stability as a function of neighborhood overlap across consecutive re-trained model instances. 
We measure the quality of the embedding by looking at a subsample of words in the training dataset, as well as words that fall in their local neighborhoods. 
We sample first naively, by picking a random subset of words, and then more strategically, by picking words that occur with high frequency, the idea being that, these words and their neighborhoods would be more accurately represented in high dimensional space. 

We observe that embedding continuous word representations is highly unstable. 
Even though average neighborhood overlap is initially (at lower neighborhood sizes) higher, the variation is much higher as well. 
As we increase the size of the neighborhood, or improve the quality of our sample by only picking the most frequent words, we do observe a reduction in embedding instability. 
However, the amount of improvement that we get is very small relative to the increase in sample size and sample quality. 
Finally, even after the embedding quality stabilizes as we consider higher neighborhood sizes, it becomes unlikely to change at relatively low values of neighborhood overlap, an indication that continuous word representations are difficult to embed consistently.
Next, we present an approach to addressing the adverse effects of sparsely distributed word representations.

\section{Stabilizing Embeddings via Latent Words}
\label{sec:soln}
Assume that we are given word embedding models, $W^i|_{i\in [1,k]}$ where $k$ is the total number of models. 
$V^i$ denotes a vocabulary of $W^i$.
A word $l$ in $V^i$ is shown as $w^i_l$ such that $W^i(w^i_l)\rightarrow \mathbb{R}^m$, and $m$ is the size of the latent space.   
Let $d(.,.)$ represent the similarity function between two vector representations and we use the cosine similarity measure defined by $d(w^i_l,w^i_m)=\frac{{\vv{w}^i_l}\cdot{\vv{w}^i_m}}{||w^i_l|||w^i_m|}$. 
Since, input co-occurrence frequencies are normalized, we expect similar results if we used Euclidean distance instead. 
Let $n^i_{\epsilon}|_{w^i_l}$ be a set of words that fall in the $\epsilon$-neighborhood of a word $w^i_l$, such that $n^i_{\epsilon} |_{{w^i_l}} =  \{  w_m^i |  d(w_m^i, w_l^i) < \epsilon, w^i_l, w^i_m \in V^i \}$. 
Our goal is to find a lower-dimensional joint subspace of these models by using common words and their neighborhoods for given models since these neighborhoods of common words are composed of polysemic and semantically drifted words. 

We first assume that the word embedding representation lies on an underlying manifold and that this manifold is locally continuous, linear and smooth. 
We then leverage the property of continuous word models to express linguistic relationships via simple linear operations. As illustrated in~\cite{mikolov2013a}, if $x, y \mbox{ and }z $ are vector representations of \textit{king}, \textit{woman}, and \textit{man}, respectively, then \textit{queen} can be extrapolated by a simple linear combination of these vectors $x+y-z$. 
In addition to analogies, other linguistic inductive tasks such as \textit{series completion} and \textit{classification} can also be solved with vector operations on word embeddings~\cite{hashimoto2016a,lee2015a}.
Inspired by these insights, we surmise that words can be added and/or subtracted to recover unobserved analogies, series completions and classifications, and as a result, to recover unobserved words. 
Even though, we don't have the exact generative model for semantically similar unobserved words, we assume they occur nearby the observed words and this gives us a mechanism for densifying the original local neighborhoods. 
Note that for the purposes of our learning objective, we don't need to make the case that the latent words are real words; we only need to ensure that they are geometrically close to where we would expect the real words to be placed. We ensure this property by construction: 
a latent word is generated by linear vector operations of original words within an $\epsilon$-neighborhood and is only included if it falls within this
neighborhood.
A latent word $w^i_{*}$ can be generated by $w^i_{*}= \sum (\alpha _n \times w^i_{r_n})$, where $\alpha _{n}$ is a randomly chosen integer from $[-1,+1]$ and  $w^i_{r_n} \in  n^i_{\epsilon}|_{w^i_l }$. 
Note that $w^i_{*}$ is a valid latent word if and only if $d(w^i_l,w^i_{*}) < \epsilon$.   
These latent points are only leveraged as anchor points to improve dimensionality reduction and alignment of the original embedding models.

\begin{figure}
  \begin{minipage}[c]{0.47\textwidth}
    \caption{Trustworthiness ($\mathbb{T}$) and Continuity ($\mathbb{C}$) for common words in $n^i_{\epsilon}|_{w^i_l}$ and $n^j_{\epsilon}|_{w^j_l}$ after finding a joint union manifold and then mapping to $R^{50}$ (larger numbers are better).}
     \label{fig:tclra}
  \end{minipage}\hfill
  \begin{minipage}[c]{0.50\textwidth}
    \includegraphics[width=0.85\textwidth]{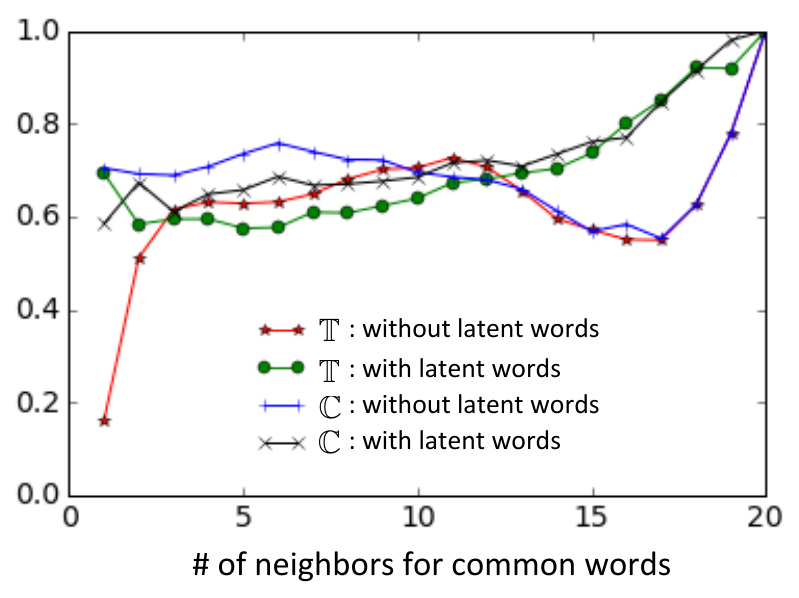}
  \end{minipage}
  \vspace{-0.90cm}
\end{figure}

Following, we present preliminary results for aligning a neighborhood of two models, $W^i$ and $W^j$, generated by the same input data and training parameters explained in Section~\ref{sec:problem} for $|V|=266K$. 
Since, we are performing manifold alignment, we illustrate via a successful alignment technique, the low rank alignment \textsc{lra} given in~\cite{boucher2015a}, which is an extension of \textsc{lle}. 
We also consider two relevant metrics to help us evaluate the quality of lower-dimensional embeddings: (i) trustworthiness ($\mathbb{T}$) and (ii) continuity ($\mathbb{C}$)~\cite{maaten2008a}. 
We could think of $\mathbb{T}$ and $\mathbb{C}$ as measuring components of the symmetric difference between a neighborhood in high and low dimensional space. In that sense, they emphasize inconsistencies with preserving the neighborhood structure in embedded space (unlike the neighborhood overlap which emphasizes the portion of the structure which is preserved). 
$\mathbb{T}$ and $\mathbb{C}$ values are analyzed for $n^i_{\epsilon}|_{w^i_l} \cap n^j_{\epsilon}|_{w^j_l}$ as presented in Fig.~\ref{fig:tclra}. 
As seen in this figure, these metrics are slightly better for our approach for up to $12$ neighbors. 
Beyond that point, the addition of latent words causes considerable improvement in the alignment performance. 
In Fig.~\ref{fig:tclra} we have illustrated alignment of local neighborhoods in $W^i$ and $W^j$ using the most common word, but we have observed similar trends when we pick less frequent words, although the larger the frequency, the larger the improvement.

In conclusion, our tailored manifold alignment approach offers a platform for fusing different word embedding models and generating richer semantic representations. Furthermore, a common representation of different models allows for explicit comparison and evaluation of the quality of the representation. In this paper, we only provide empirical results for alignment of a selected subset of local neighborhoods. As a future work, we plan (i) to extend our approach to generate a holistic embedding model that optimizes alignment across all local neighborhoods, (ii) to add more constraints into $\textsc{lra}$ process such that weights of latent words in the similarity matrix will be a part of loss function to minimize and (iii) to characterize the effects of anchor points in different models.

\bibliography{vec2vecSahin,vec2vecCaceres}
\bibliographystyle{iclr2017_workshop}

\end{document}